\documentclass[10pt,twocolumn,letterpaper]{article}

\usepackage[pagenumbers]{cvpr} %

\usepackage{graphicx}
\usepackage{amsmath}
\usepackage{amssymb}
\usepackage{booktabs}
\PassOptionsToPackage{numbers, compress}{natbib}
\usepackage{natbib}
\usepackage{enumitem}
\usepackage{symbols}
\usepackage{wrapfig}
\usepackage{textcomp}
\usepackage{hyperref}
\hypersetup{pagebackref,breaklinks,colorlinks}

\usepackage{color}
\usepackage{xcolor}
\definecolor{mycitecolor}{rgb}{0.08984375, 0.41796875, 0.93359375} %

\usepackage[utf8]{inputenc} %
\usepackage[T1]{fontenc}    %
\usepackage{nicefrac}       %
\usepackage{microtype}      %
\usepackage{pmboxdraw}
\usepackage{graphicx}
\usepackage{comment}
\usepackage{symbols}
\usepackage{multirow}
\usepackage{ctable} %
\usepackage{pifont}%
\newcommand{\xmark}{\ding{55}}%
\usepackage[ruled,linesnumbered]{algorithm2e} 
\SetKwComment{Comment}{$\triangleright$\ }{}

\usepackage{amsmath,amsfonts,bm}

\def\eqref#1{~\ref{#1}}

\def\1{\bm{1}}

\def\sp{space}

\def\vtheta{{\bm{\theta}}}
\def\vphi{{\bm{\phi}}}
\def\va{{\bm{a}}}

\def\vx{{\bm{x}}}

\def\mW{{\bm{W}}}

\DeclareMathAlphabet{\mathsfit}{\encodingdefault}{\sfdefault}{m}{sl}
\SetMathAlphabet{\mathsfit}{bold}{\encodingdefault}{\sfdefault}{bx}{n}

\def\gA{{\mathcal{A}}}

\def\gD{{\mathcal{D}}}
\def\gE{{\mathcal{E}}}
\def\gF{{\mathcal{F}}}

\def\gL{{\mathcal{L}}}

\def\gV{{\mathcal{V}}}

\def\sR{{\mathbb{R}}}

\newcommand*{\ShowNotes}{} %

\usepackage{listings}
\DeclareFixedFont{\ttb}{T1}{txtt}{bx}{n}{9} %
\DeclareFixedFont{\ttm}{T1}{txtt}{m}{n}{9}  %

\usepackage{color}
\definecolor{deepblue}{rgb}{0,0,0.5}
\definecolor{deepred}{rgb}{0.6,0,0}
\definecolor{deepgreen}{rgb}{0,0.5,0}
\lstnewenvironment{python}[1][]
{
\pythonstyle
\lstset{#1}
}
{}

\newcommand\pythoninline[1]{{\pythonstyle\lstinline!#1!}}

\newcommand\pythonstyle{\lstset{
language=Python,
basicstyle=\ttm,
morekeywords={self},              %
keywordstyle=\ttb\color{deepblue},
emph={MyClass,__init__},          %
emphstyle=\ttb\color{deepred},    %
stringstyle=\color{deepgreen},
frame=tb,                         %
showstringspaces=false
}}

\newcommand{\ourname}[0]{EH-DNAS}

\def\cvprPaperID{6576} %
\def\confName{CVPR}
\def\confYear{2022}

\begin{document}
\definecolor{darkred}{rgb}{0.7,0.1,0.1}
\definecolor{darkgreen}{rgb}{0.1,0.7,0.1}
\definecolor{cyan}{rgb}{0.7,0.0,0.7}
\definecolor{dblue}{rgb}{0.2,0.2,0.8}
\definecolor{maroon}{rgb}{0.76,.13,.28}
\definecolor{burntorange}{rgb}{0.81,.33,0}
\definecolor{tealblue}{rgb}{0.212,0.459, 0.533}

\definecolor{pp}{rgb}{0.43921569, 0.18823529, 0.62745098}
\definecolor{rr}{rgb}{0.5254902 , 0.00784314, 0.12941176}
\definecolor{bb}{rgb}{0.09019608, 0.23529412, 0.37647059}
\definecolor{yy}{rgb}{0.49803922, 0.3372549 , 0.0}
\definecolor{gg}{rgb}{0.02352941, 0.3372549 , 0.17647059}

\ifdefined\ShowNotes
  \newcommand{\colornote}[3]{{\color{#1}\bf{#2: #3}\normalfont}}
\else
  \newcommand{\colornote}[3]{}
\fi

\newcommand {\minh}[1]{\colornote{blue}{Minh}{#1}}
\newcommand {\minhw}[1]{{\color{blue}{#1}}}

\newcommand {\ray}[1]{\colornote{tealblue}{Ray}{#1}}
\newcommand {\rayw}[1]{{\color{blue}{#1}}}

\newcommand {\xz}[1]{\colornote{burntorange}{XiaoFan}{#1}}
\newcommand {\xzw}[1]{{\color{gg}{#1}}}

\newcommand {\qian}[1]{\colornote{maroon}{Qian}{#1}} %
\newcommand {\qianw}[1]{{\color{rr}{#1}}}  %

\newcommand{\eat}[1]{} %

\title{{\ourname}: End-to-End Hardware-aware\\ Differentiable Neural Architecture Search \vspace{-0.3cm}}

\author{Qian Jiang\thanks{indicates equal contribution.}~$^1$, ~Xiaofan Zhang\footnotemark[1]~$^1$, ~Deming Chen$^1$, ~Minh N. Do$^1$, ~Raymond A. Yeh$^2$\\
$^1$University of Illinois at Urbana-Champaign \\
$^2$Toyota Technological Institute at Chicago\\
{\tt\small \{qianj3, xiaofan3, dchen, minhdo\}@illinois.edu \qquad yehr@ttic.edu}
}
\maketitle

\begin{abstract}
\vspace{-0.3cm}
In hardware-aware Differentiable Neural Architecture Search (DNAS), it is challenging to compute gradients of hardware metrics to perform architecture search. Existing works rely on linear approximations with limited support to customized hardware accelerators. In this work, we propose End-to-end Hardware-aware DNAS (EH-DNAS), a seamless integration of end-to-end hardware benchmarking, and fully automated DNAS to deliver hardware-efficient deep neural networks on various platforms, including Edge GPUs, Edge TPUs, Mobile CPUs, and customized accelerators. 
Given a desired hardware platform, we propose to learn a differentiable model predicting the end-to-end hardware performance of neural network architectures for DNAS. 
We also introduce E2E-Perf, an end-to-end hardware benchmarking tool for customized accelerators. 
Experiments on CIFAR10~\cite{krizhevsky2009learning} and ImageNet~\cite{ILSVRC15} show that EH-DNAS improves the hardware performance by an average of $1.4\times$ on customized accelerators and $1.6\times$ on existing hardware processors while maintaining the classification accuracy. 
\vspace{-0.3cm}
\end{abstract}

\section{Introduction}
Deep neural networks (deep-nets) have achieved numerous successes in computer vision~\cite{lecun2015deep, kaiming2016resnet, sze2015inception, vgg, ash2017attention}. The design of architecture plays a crucial role for deep-nets. Hence, neural architecture search (NAS)~\cite{zoph2017neural,zoph2018learning,liu2018progressive, tan2019mnasnet,pham2018efficient,ying2019bench} has become significantly important. 

Earlier works~\citep{zoph2017neural,baker2017designing} have focused on improving the model's prediction accuracy. Recent approaches have additionally considered hardware-related metrics, \eg, model inference latency, to further enable %
practical deep-net designs on mobile and embedded platforms~\citep{wu2019fbnet, tan2019mnasnet, hao2019fpga, stamoulis2019single, zhang2020skynet, jiang2020standing}.

Existing hardware-aware NAS methods rely on proxies to approximate the end-to-end hardware performance of network candidates. A few common proxies are the number of multiply-accumulate operations and network parameters~\citep{liu2019darts,real2019regularized}. %
However, these proxy metrics may fail to align with the realistic hardware performance as they may be highly unrelated. For example, deep-nets with fewer operations or memory footprints do not necessarily lead to higher hardware efficiency or performance~\citep{tan2019mnasnet}. 

To address this gap, existing works~\cite{ChamNet2019,cai2019proxylessnas, wu2019fbnet} adopt linear approximations of hardware metrics. In other words, look-up-tables (LUTs) %
consist of per-layer hardware metrics. %
The overall network's hardware metric is simply estimated by a weighted summation of all the layers.

Another solution is to incorporate the end-to-end hardware performance, such as using the latency and throughput of running an entire network model on targeted devices~\citep{zhang2020skynet,dudziak2020brp}. However, this approach can not be directly adopted by the latest DNAS procedures due to its discrete measurements of each architecture. 
Additionally, the hardware feedback is hard to obtain in an online fashion during NAS, as it requires additional time and effort to evaluate the hardware performance. Support for customized accelerators and dedicated hardware architectures may also be lacking.

In this work, we address these challenges by proposing {\ourname}, an end-to-end hardware-aware DNAS framework that automatically searches for hardware-efficient deep-nets on various types of hardware. 
Specifically, we propose to learn a differentiable function to approximate the hardware performance on a given hardware platform. Therefore, this function provides hardware feedback that can be directly integrated into the DNAS flow, enabling the search for efficient and accurate architectures. We also provide E2E-Perf, a benchmarking tool, to benchmark hardware performance on customized hardware accelerators.

We evaluate our approach on CIFAR10 and ImageNet using the search space of DARTS~\cite{liu2019darts} and HW-Bench-201~\cite{hwnasbench2021,dong2021nats}. We show that EH-DNAS improves the hardware performance by an average of 1.4$\times$ on customized accelerators, and 1.6$\times$ on existing hardware processors, including Edge GPUs, Edge TPUs, and Mobile CPUs, while maintaining the classification accuracy. 

{\bf \noindent Our contributions:}
\begin{itemize}[leftmargin=0.6cm]
    \itemsep0em 
    \item We propose {\ourname} to integrate hardware metrics benchmarking, differentiable hardware loss approximation, and DNAS to search for hardware-aware network architectures.
    \item We provide E2E-Perf to efficiently and accurately benchmark hardware metrics of customized hardware accelerators.
    \item We empirically show on CIFAR10 and ImageNet that {\ourname} improves hardware performance over SOTA on various hardware platforms. %
\end{itemize}

\begin{figure*}[t]
    \centering
    \vspace{-0.2cm}
    \includegraphics[width=\textwidth]{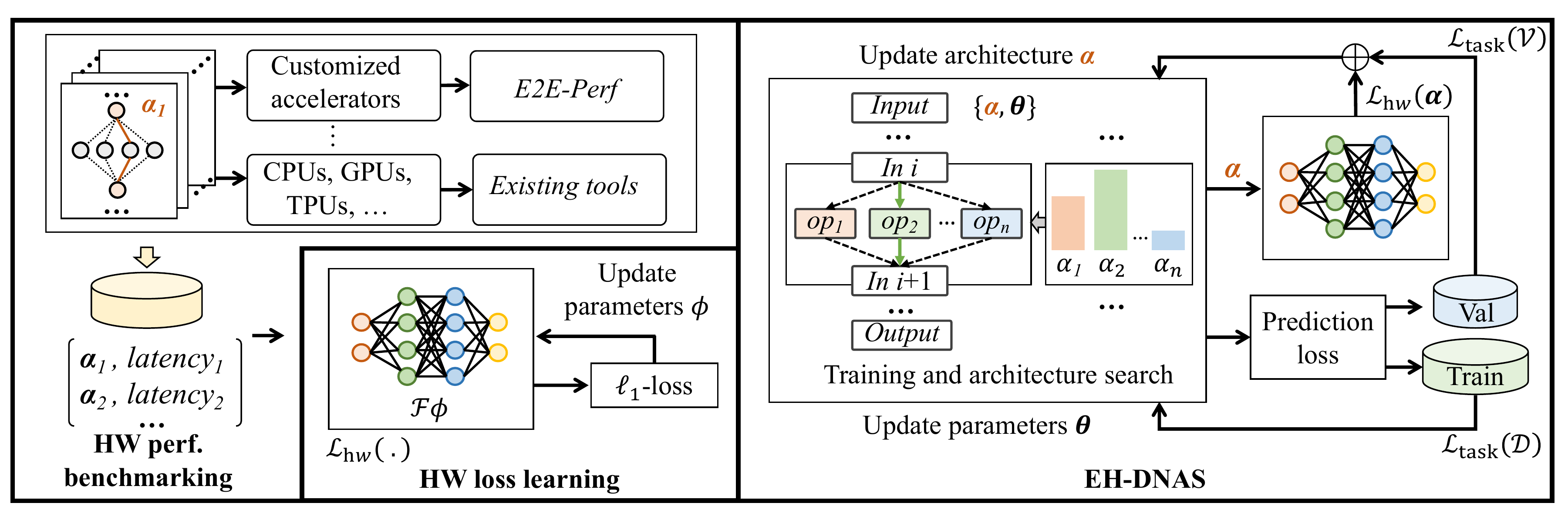}
    \vspace{-0.7cm}
    \caption{Illustration of our proposed {\ourname} framework consisting of three stages. {\bf HW performance benchmarking:} We measure the hardware (HW) performance of network architectures $\{\va_i\}_{i=1}^n$ to collect an HW performance dataset $\{\va_i, {latency}_i\}_{i=1}^n$. Specifically, we propose E2E-Perf to benchmark customized accelerators. This stage enables diverse hardware support for {\ourname}.
    {\bf HW loss learning:} With a collected dataset, we propose a deep-net to learn a differentiable approximation $\gF_\phi$ of the hardware performance with respect to any selected network architecture. 
    {\bf EH-DNAS:} We then perform DNAS with learned HW loss providing end-to-end hardware feedback, \ie, $\gL_{\tt hw}(\va) \triangleq \gF_\phi(\va)$.
    }
    \label{fig:pipeline}
    \vspace{-0.45cm}
\end{figure*}

\section{Related Work}

Proxy-based hardware estimation is a widely adopted method to provide hardware benchmarking for NAS designs. However, earlier works have indicated that the popular proxies,~\ie, FLOPS and network parameter number, are not closely correlated with actual inference speed running on hardware~\cite{sandler2018mobilenetv2,wu2019fbnet,wan2020fbnetv2}. To improve, recent works use look-up-tables~(LUTs)~\cite{ChamNet2019,wu2019fbnet,hwnasbench2021} and end-to-end evaluation tools~\cite{zhang2020skynet,dudziak2020brp} for hardware-aware NAS designs. These methods can collect more realistic hardware metrics by running deep-nets on the targeted accelerators. Proxyless NAS~\cite{cai2019proxylessnas} models per operator latency and use linear approximation for end-to-end performance.

However, these linear approximation methods may fail to accurately represent end-to-end hardware metrics, and they also lack support for customized accelerators with varied architectures and optimization strategies. This is because the hardware overheads between major operations (\eg, network layers, blocks) and customized optimizations tend to be neither independent nor linear across layers.
For example, 
when targeting the same VGG network~\cite{vgg} with identical hardware resource budgets, the throughput performance of accelerator designs~\cite{qiu2016going, xiao2017exploring, zhang2018dnnbuilder} varies from 137 to 262 giga-operation per second (GOPS). The performance variation is mainly caused by the different architecture-level compute and memory access patterns. These differences are hard to be captured by linear approaches.
Additionally, even for the same architecture paradigm, an optimization called column-based cache proposed by early work~\cite{zhang2018dnnbuilder} can bring 7.7$\times$ and 43$\times$ improvements regarding latency and memory footprint. 
This type of optimization cannot be represented with LUTs as it introduces a more fine-grained design that mainly affects the accelerator's end-to-end performance. 

While the end-to-end methods show better consistency for benchmarking the end-to-end performance, they have some shortcomings, \eg, they need extra time and effort for measurements on actual hardware and cannot be integrated into existing DNAS procedures due to the discrete nature. An Electronic Design Automation (EDA) toolset~\cite{xilinx_hls} was adopted to collect end-to-end hardware performance following the process of high-level synthesis~\cite{zhang2020skynet}. It requires hours for getting hardware feedback which is inefficient to be integrated into recent NAS methods. Another work~\cite{dudziak2020brp} adopts a graph convolutional network (GCN) for end-to-end latency prediction by encoding the architecture as an adjacency matrix. Yet the predictor is non-differentiable since it is unclear how to perform backpropagation through the adjacency matrix. Hence, it cannot be directly integrated into the recently popular end-to-end trainable DNAS flows.

\section{Preliminaries}\label{sec:prelim}
We provide a review on differentiable neural architecture search, \eg, (DNAS)~\citep{liu2019darts}, and its application to search for models that run efficiently on hardware~\citep{wu2019fbnet}.

{\noindent \bf Differentiable Neural Architecture Search.}
At a high level, neural architecture search~(NAS) can be formulated as a bi-level optimization problem:
\be\label{eq:dnas}
\min_{\va \in \gA} \cL(\gV, \vtheta^*(\va), \va) \;\; \text{s.t}\;\; \vtheta^*(\va)=\min_{\vtheta}\cL(\gD, \vtheta, \va),
\ee
where $\va$ denotes the parameters representing an architecture, $\cL$ denotes a loss function, $\gV$ denotes the validation dataset, $\gD$ denotes the training dataset, $\theta$ denotes the model's trainable parameters, and $\gA$ denotes the architecture search space. The goal is to find an architecture $\va$, within the search space $\gA$, to achieve the lowest $\cL$ on the dataset $\gV$ when $\theta$ is trained on $\gD$. 

The search space consists of $L$ layers where each layer consists of $K$ candidate blocks, \eg, convolution with different filter sizes. Formally, an architecture $\va$ and the search space $\gA$ is defined as follows:
\be
\va = [a_1^{(1)}, \hdots a_k^{(l)},\hdots a_K^{(L)}] \text{ and } \gA = \{0,1\}^{K \times L},
\ee
where $a_k^{(l)} \in \{0,1\}$ and $\sum_k a_k^{(l)}=1$ indicate the selection of a $f_k^{(l)}$ denoting the $k^{\text{th}}$ block at the $l^{\text{th}}$ layer.

Brute-force search over the discrete space $\gA$ is computationally expensive. Hence, DNAS relaxes $a_k^{(l)} \in [0,1]$ using a soft-max reparameterization. The selection of a block is formulated as a weighted sum. A layer $l$'s output is defined as
\be
\vx^{(l+1)}=\sum_{k} a_k^{(l)}\cdot f_k^{(l)}(\vx^{(l)};\theta).
\ee
With these two modifications, the architecture is now differentiable with respect to (\wrt) the loss function. This enables the use of gradient-based optimization techniques to perform architecture search.
Typically, model weights $\vtheta$ with training loss and architecture selection parameter $\va$ with validation loss are alternatively updated to approximate the solution to the bi-level optimization in~\equref{eq:dnas}. 

{\noindent \bf Hardware-Aware Search.} To utilize DNAS for designing hardware efficient architecture, it is typical to choose a loss function of the form
\be\label{eq:ltask_lhardware}
\cL(\cV, \vtheta, \va) = \cL_{\tt task}(\cV, \vtheta, \va) + \beta\cL_{\tt hw}(\va),
\ee
where $\beta \in \sR^+$ is a hyperparameter that balances the two losses. Here, $\cL_{\tt task}$ captures the task performance, \eg, cross-entropy for classification tasks and $\cL_{\tt hw}$ captures the hardware cost of an architecture, \eg, latency.
 For example, FBNet~\cite{wu2019fbnet} penalizes an architecture by the sum of latency of each block, 
 \be\label{eq:fb_net_hardware}
 \cL_{\tt hw}^{\tt LUT}(\va) \triangleq \sum_l \sum_k a_k^{(l)} \cdot \text{LAT}(f_k^{(l)}),
 \ee
 where LAT is a lookup table (LUT) that records the latency of block $f_k^{(l)}$. The LUT-based methods can only record hardware metrics of major network operations (\eg, blocks, layers) but fail to capture overheads between operations. Inevitably, these methods will ignore particular hardware costs, such as data pre-/post-processing and data access latency, and cause inaccurate benchmarking. Also, architecture-level optimizations cannot be accurately reflected using the LUTs.

\section{{\ourname} for Hardware Aware NAS}
Our goal is to effectively search over hardware-aware models. As reviewed in~\secref{sec:prelim}, the search for a hardware-aware model is based on a hardware loss function $\cL_{\tt hw}$. Given an architecture $\va$, $\cL_{\tt hw}(\va)$ should accurately characterize the hardware's metrics, \eg, latency. At the same time, $\cL_{\tt hw}$ needs to be computed efficiently and differentiable \wrt the architecture $\va$.
To accomplish these goals, we propose {\ourname} to: (a) learn a differentiable approximation of the hardware feedback (Sec.~\ref{sec:hw_loss}) to be integrated into differentiable architecture search;
(b) adopt proper tools for measuring/benchmarking realistic hardware performance (Sec.~\ref{sec:cap_hw_perf}). We provide a visual overview of our {\ourname}  in~\figref{fig:pipeline}.

\subsection{Learning a Differentiable Hardware Loss}
\label{sec:hw_loss}
A hardware loss, $\gL_{\tt hw}(\va)$ should accurately resemble the desired hardware metrics, \eg, latency, for a given model architecture $\va$. To use the hardware loss with DNAS, we also need $\gL_{\tt hw}(\va)$ to be differentiable \wrt $\va$. To accomplish this, we propose to learn this loss function parameterized with a deep-net.

We formulate this learning procedure as a regression task:
\be\label{eq:e2e-prof-approx}
\min_{\vphi} \frac{1}{|\gA|}\sum_{\va \in A}|\text{HW Perf.}(\va)-\gL_{\tt hw}^{\tt deep}(\va;\phi)|,
\ee
to minimize the mean absolute error between the hardware performance $\text{HW Perf.}(\va)$ and the hardware loss $\gL_{\tt hw}^{\tt deep}$ over all architectures in $\gA$.
We describe tools to acquire $\text{HW Perf.}(\va)$ in~\secref{sec:cap_hw_perf}.
We parameterize $\gL_{\tt hw}^{\tt deep}$ using a deep-net with trainable parameters $\phi$, \ie,
\bea
\gL_{\tt hw}^{\tt deep} (\va; \phi) \triangleq \gF(\va;\phi) =  g_N \circ g_{N-1} \hdots \circ g_1(\va),
\eea
where $\phi$ subsumes all the trainable parameters in the layers $g_N$ to $g_1$.

In more detail, the deep-net $\gF$ starts with a linear embedding layer $g_1$, which represents each candidate block $a_k^{(l)}$, the $k$-th candidate block at $l$-th layer, with a 10-dimensions vector. Formally, an architecture's embedding $\gE(\va)$ is computed as:
\be
\gE(\va) = \text{Concat}([\mW_1 \va^{(1)}, \hdots, \mW_L\va^{(L)} ]),
\ee
where 
$\va^{(l)} = [a_1^{(l)} \hdots, a_K^{(l)}]$,
and $\mW_l$ denotes the trainable parameters of the embedding layer. Following the embedding layer are three fully connected layers with ReLU non-linearities. We apply dropout~\cite{Srivastava2014DropoutAS} to the final layer as we observed over-fitting to the training set. 

To train this model, we need to compute~\equref{eq:e2e-prof-approx} and use gradient methods as the model involves a deep-net. However, this is not always feasible due to the size of the search space $|\gA|$. Hence, when the size of $|\gA|$ is enormous, we uniformly sample random architectures from $\gA$ and update the model parameters using mini-batch gradient descent. 

With a $\gL_{\tt hw}^{\tt deep}$ trained, we can easily compute an approximation of architecture's hardware performance by running a forward pass through the deep-net. The gradient \wrt the architecture can also be computed by running a backward pass through the deep-net, \ie,
\bea
\frac{\partial \gL_{\tt hw}^{\tt deep} (\va)}{\partial \va} = \frac{\partial g_N}{\partial g_{N-1}} \cdot
\frac{\partial g_{N-1}}{\partial g_{N-2}} \hdots \frac{\partial g_1}{\partial \va}.
\eea
Hence, we can easily integrate $\gL_{\tt hw}^{\tt deep}$ into any DNAS method to perform hardware aware DNAS and search for hardware efficient models.
The benefits of our deep hardware loss $\gL_{\tt hw}^{\tt deep}$ are that it does not assume independence nor linearity among the candidate blocks. Hence, it more realistically captures the hardware metrics than the LUT approach reviewed in~\equref{eq:fb_net_hardware}, which can only capture additive behavior. 

We can apply our approach for hardware metrics from various hardware platforms,~\eg, Edge GPUs, Edge TPUs, and Mobile CPUs.
These existing hardware designs may not always be suitable for handling emerging deep-net models, especially those with strict customization requirements. Customized hardware accelerators are necessary as their architectures and configurations are highly domain-specific. It means they are more powerful and efficient, which can perfectly fit the edge deep-net deployment. Although important, we found that the support for accurately and efficiently benchmarking hardware performance on customized hardware accelerators for NAS is lacking. Hence, we propose E2E-Perf which we describe next.

\begin{figure}
    \centering
    \vspace{-0.16cm}
    \includegraphics[width=\columnwidth]{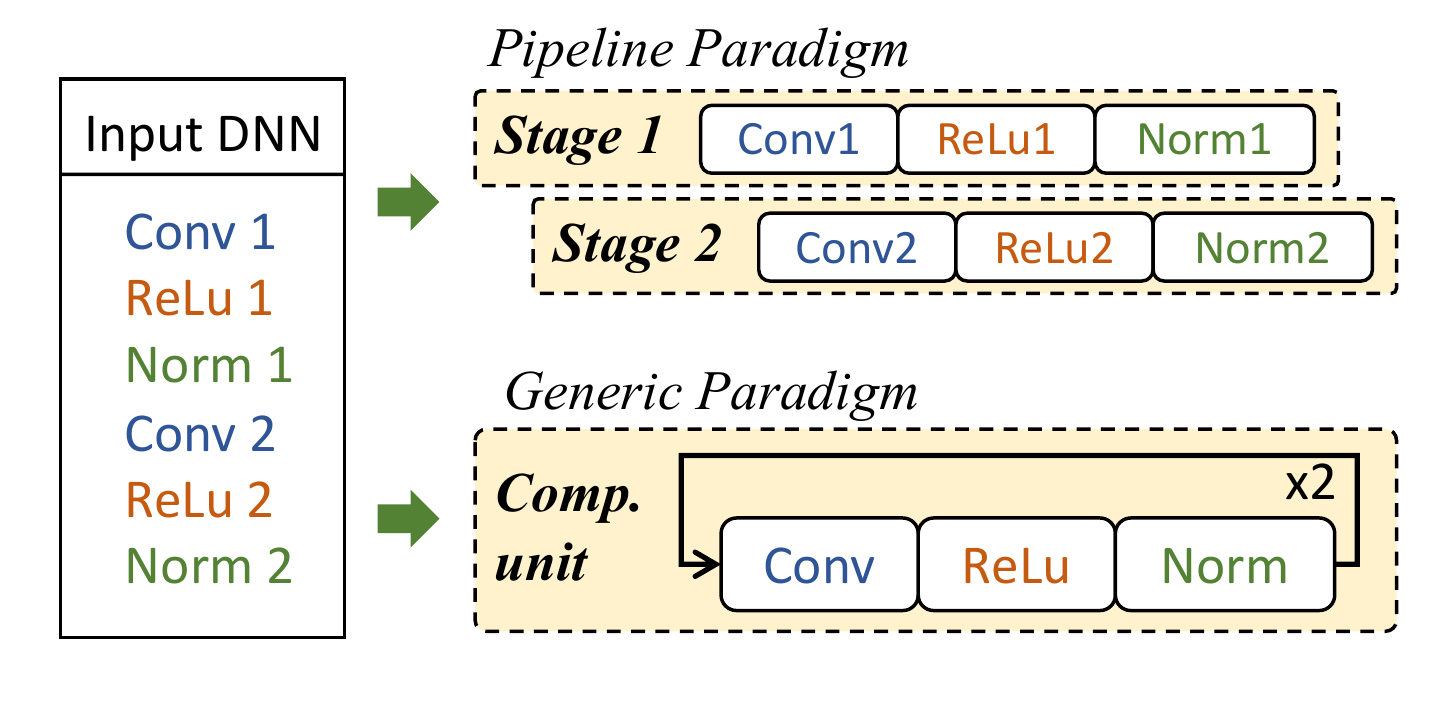}
    \vspace{-0.8cm}
    \caption{Different paradigms cause different hardware execution patterns even for running the same network.}
    \label{fig:arch_paradigm}
    \vspace{-0.5cm}
\end{figure}

\subsection{Benchmarking Hardware Performance}
\label{sec:cap_hw_perf}

{\bf\noindent For customized hardware accelerators.}
Customized accelerators are actively developed to provide improved performance and efficiency for deep-nets. To obtain their performance feedback, we propose E2E-Perf to perform accurate end-to-end benchmarking for customized deep-net accelerators.
Compared to existing tools, E2E-Perf is fully automated with direct support to most deep-nets under any hardware budget. It can generate end-to-end hardware metrics instantly after taking an architecture and arbitrary hardware budget, so there is no need to perform manual data collection. It supports customized architecture paradigms. It also comes with a design space exploration engine to optimize hardware configuration following various paradigm-specific optimization strategies.

The proposed E2E-Perf contains three stages: 1) network model analysis, 2) customized architecture modeling, and 3) paradigm-specific optimizations, to provide hardware feedback given the input network candidate and hardware resource budgets. 
In the first stage, the network definition files and available hardware budgets are passed to E2E-Perf for network model analysis and resource boundary setup. 
Next, a particular architecture paradigm is selected to continue benchmarking. E2E-Perf supports two popular customized accelerator paradigms, including the pipeline paradigm \citep{zhang2018dnnbuilder,wei2018tgpa} and the generic paradigm \citep{chen2016eyeriss,jouppi2017tpu,ye2020hybriddnn} (Fig. \ref{fig:arch_paradigm}). These two paradigms come with their unique optimization opportunities that lead to significantly different hardware design spaces, compute and memory access patterns, resulting in various achievable hardware performance and costs.
In the last stage, E2E-Perf explores the accelerator design spaces and provides paradigm-specific optimization strategies given the input constraints. More details of E2E-Perf are included in the Appendix.

\begin{table}[t]
    \centering
\setlength{\tabcolsep}{2pt}
\small
\resizebox{0.99\linewidth}{!}{
    \begin{tabular}{cccc}
    \specialrule{.15em}{.05em}{.05em}
      Estimator & Avg. Error & Min Error & Max Error \\
      \hline
      \hline
        AutoDNNchip \cite{xu2020autodnnchip} & 5.20\% & 2.12\% &  7.67\% \\
        HybridDNN \cite{ye2020hybriddnn} & 4.03\% & - & - \\
        DNN-chip predictor \cite{zhao2020dnn} & - & - & 16.84\% \\ \hline
        E2E-Perf (PP) & 1.15\% &	0.18\% & 2.26\% \\
        E2E-Perf (GP) & 2.17\% & 0.29\% & 8.65\%  \\
    \specialrule{.15em}{.05em}{.05em}     
    \end{tabular}
}    
\vspace{-0.2cm}
    \caption{Estimation errors between the estimated and the board-level performance of customized deep-net accelerators. (PP) and (GP) denote the pipeline paradigm and the generic paradigm adopted by E2E-Perf.}
    \label{tab:e2e_perf_err}
    \vspace{-0.5cm}
\end{table}
\begin{table*}[t]
    \centering
    \small
    \setlength{\tabcolsep}{7pt}
    \begin{tabular}{ccccccccc}
    \specialrule{.15em}{.05em}{.05em}
      \multirow{2}{4em}{Approach} & \multicolumn{3}{c}{Latency (ms)$\downarrow$}& Test Error$\downarrow$ & Params$\downarrow$  & \#ops & \#GFLOP$\downarrow$ & Search Cost$\downarrow$\\
      \cmidrule{2-4}
       & Small & Medium & Large &(\%) &(M)& & & (GPU Days)\\
      \hline
      \hline
        NASNet-A~\cite{tan2019mnasnet}   & \xmark & \xmark & 1.38 & 2.83 & 3.1  & 13 & 0.12 & 2000 \\
        AmoebaNet-A~\cite{real2019regularized}  & \xmark & 3.73 & 0.93 & 3.12 & 3.1 & 19 & 0.11 & 3150\\
        SNAS-mild~\cite{xie2018snas} & \xmark & 1.24 & 0.31 & 2.98 & 2.9 & 7 & 0.08 &  1.5\\
        DARTS~\cite{liu2019darts}  & \xmark & \bf 1.66 & \bf 0.55 & \bf 3.00 & \bf3.3 & 7 & \bf0.09 & 1\\

        DARTS + EH-DNAS & \bf 2.49 & \bf 0.83 & \bf 0.31 & \bf 2.84 &  \bf2.5 & 7 &  \bf0.07 & 1 \\
    \specialrule{.15em}{.05em}{.05em}     
    \end{tabular}
\vspace{-0.1cm}
    \caption{Quantitative results on CIFAR-10 on DARTS search space. We evaluate latency on customized hardware accelerators with the pipeline paradigm, using E2E-Perf. We report the latency of the searched architecture on three different hardware budgets. {\xmark} indicates the architecture exceeds the HW budget. Note that LUT can not be obtained with PP.}
    \label{tab:cifar10}
    \vspace{-0.2cm}
\end{table*}

\begin{table*}[t]
    \centering
    \small 
    \setlength{\tabcolsep}{6pt}
    \begin{tabular}{ccccccccc}
    \specialrule{.15em}{.05em}{.05em}
      \multirow{2}{4em}{Approach} & \multicolumn{3}{c}{Latency (ms)$\downarrow$}& Test Error$\downarrow$ & Params$\downarrow$  & \#ops & \#GFLOP$\downarrow$ & Search Cost$\downarrow$\\
      \cmidrule{2-4}
      & Small & Medium & Large & (\%) &(M)& & &(GPU Days)\\
      \hline
      \hline
        NASNet-A~\cite{tan2019mnasnet} & 12.82 & 7.58 & 5.49 & 2.83 & 3.1  & 13 & 0.12 & 2000 \\
        AmoebaNet-A~\cite{real2019regularized} & 9.71 & 5.88 & 4.39 & 3.12 & 3.1 & 19 & 0.11 & 3150\\
        SNAS-mild~\cite{xie2018snas} & 4.59 & 2.97 & 2.48 & 2.98 & 2.9 & 7 & 0.08 & 1.5\\
        DARTS~\cite{liu2019darts}  & \bf5.59 & \bf3.60 & \bf2.99 & \bf3.00 & \bf3.3 & 7 & \bf0.09 & 1\\
        DARTS + FBNet LUT~\cite{wu2019fbnet} & \bf5.99 & \bf3.72 & \bf2.93 & \bf2.82 & \bf2.9 & 7 & \bf0.08 & 1\\
        DARTS + EH-DNAS & \bf4.41 & \bf2.82 & \bf2.31 & \bf2.92 & \bf2.5 & 7 & \bf0.07 & 1\\
    \specialrule{.15em}{.05em}{.05em}     
    \end{tabular}
    \vspace{-0.1cm}
    \caption{Quantitative results on CIFAR-10 on DARTS search space. We evaluate latency on customized hardware accelerators with the generic paradigm, using E2E-Perf. We report the latency of the searched architecture on three different hardware budgets. }
    \label{tab:cifar10_p2}
    \vspace{-0.4cm}
\end{table*}

To validate the proposed E2E-Perf, we compare the estimated hardware performance of the customized accelerators to their measured results from FPGA board-level implementation. As shown in Table \ref{tab:e2e_perf_err}, the estimation error introduced by E2E-Perf is 1.15\% on average (range 0.18\% to 2.26\%) for the pipeline paradigm and 2.17\% on average (range 0.29\% to 8.65\%) for the generic paradigm. 
Compared to the recently published tools \cite{xu2020autodnnchip, ye2020hybriddnn,zhao2020dnn}, E2E-Perf provides more accurate performance estimation and significantly improves the hardware feedback quality for guiding the network architecture search in {\ourname}.

{\bf\noindent For existing hardware processors.}
{\ourname} can also support the existing hardware processors benchmarking tool for hardware performance feedback. These tools,~\eg, HW-NAS-Bench \cite{hwnasbench2021}, collect the measured/estimated hardware performance of all the networks in the search spaces of NAS-Bench-201~\cite{dong2020nasbench201} on several hardware devices. HW-NAS-Bench provides our desired hardware performance datasets for training differentiable hardware losses.
In this paper, we adopt HW-NAS-Bench \cite{hwnasbench2021} to show the effectiveness of {\ourname} for four hardware devices, including Edge GPU (NVIDIA TX2), Edge TPU, and two Mobile CPUs (on Raspberry Pi 4 and Pixel 3 mobile phone).
\subsection{Architecture Search Details for {\ourname}}
\label{sec:e2e_search}
The learned differentiable hardware loss $\gL_{\tt hw}^{\tt deep}$ is trained to predict the end-to-end latency of the network. Hence, the overall validation loss function for Hardware-aware DNAS is
\bea
\cL(\cV, \vtheta, \va) = \cL_{\tt task}(\cV, \vtheta, \va) + \beta\cL_{\tt hw}^{\tt deep}(\va; \phi)
\eea
with $\phi$ fixed during the architecture search.
In our experiments, we demonstrate the effectiveness of our hardware loss following the DNAS setup in DARTS V1~\cite{liu2019darts} and NAS-Bench-201~\cite{dong2020nasbench201},~\eg, the procedure for searching architecture, training searched architecture from scratch and evaluation. %
Please see supplementary materials for details. %

For hyperparameters, we tune $\beta$ in~\equref{eq:ltask_lhardware}, controlling the scale of the hardware loss term, using grid search over the range $\{ 0.1, 0.01, 0.005, 0.001, 0.0005, 0.0001\}$. Other hyperparameters follow DARTS and NAS-Bench201's default.

\section{Experiments}\label{sec:exp}
We evaluate the proposed approach in three folds. First, we evaluate the performance of our framework on DARTS~\cite{liu2019darts} search space, where we focus on customized hardware accelerator performance using our proposed tool E2E-Perf. We evaluate searched architectures on CIFAR10 and their performance when transferring to ImageNet. Next, we evaluate on NAS-Bench-201~\cite{dong2020nasbench201, dong2021nats} search space, where we focus on existing hardware performance~(Edge GPU, Edge TPU, Raspi 4, Pixel 3) using HW-NAS-Bench~\cite{hwnasbench2021}. We evaluate the performance of searched architectures on CIFAR10. We aim to search for neural architectures with optimized hardware performance without sacrificing classification accuracy. Last, we provide analysis of our framework by a) quantifying how accurate the hardware loss resembles hardware performance; b) analyzing the effect of model size on hardware loss; c) studying the trade-off between classification accuracy and hardware performance; and d) examining the searched cells of different approaches.
\begin{table*}[h]
\small
    \centering
    \begin{tabular}{cccccccccc}
    \specialrule{.15em}{.05em}{.05em}
      \multirow{2}{4em}{Approach} & \multicolumn{3}{c}{Latency~(ms)$\downarrow$}& Top1 Error$\downarrow$ & Params$\downarrow$ & \#ops & \#GFLOP$\downarrow$ & Search Cost$\downarrow$\\
      \cmidrule{2-4}
       & Small & Medium & Large  & (\%) & (M)& & & (GPU Days)\\
      \hline
      \hline
        NASNet-A~\cite{tan2019mnasnet}  & 28.57&16.67 & 11.90 & 26.0  & 5.3  & 13 & 1.23 & 2000 \\
        AmoebaNet-A~\cite{real2019regularized} & 20.83 & 12.50 & 9.17 & 25.5 & 5.1 & 19 & 1.07 & 3150\\
        ProxylessNAS~\cite{cai2019proxylessnas} & 71.43 & 37.04 & 20.41 & 24.9 & 4.1 & 7 & 0.66 & 200\\
        MobileNet-V3~\cite{Howard_2019_mobilenetv3} & 28.57 & 15.87& 9.90 & 26.0 & 5.5 & - & 0,12 & -\\
        FBNet-A~\cite{wu2019fbnet} &38.46&20.41& 11.63 & 27.0 & 4.3 & 9 & 0.48 & 9\\
        SNAS-mild~\cite{xie2018snas} & 11.49 & 7.41& 6.02 & 27.3 & 4.3 & 7 & 0.89 & 1.5 \\
        \hline
        \addlinespace
        DARTS~\cite{liu2019darts} & \bf14.71 & \bf9.26 & \bf7.46 & \bf30.8 & \bf4.7 & 7 & \bf1.03 & 1\\

        DARTS + FBNet LUT~\cite{wu2019fbnet} & \bf 15.80 & \bf 9.68  & \bf 7.40 & \bf 30.0 & \bf 4.1 & 7 & \bf0.94 & 1\\
        DARTS + EH-DNAS &  \bf11.49&\bf 7.30 & \bf 5.81 & \bf 30.4 & \bf3.7 & 7 & \bf0.84 & 1\\
    \specialrule{.15em}{.05em}{.05em}     
    \end{tabular}
    \vspace{-0.2cm}
    \caption{Quantitative results on ImageNet on DARTS search space. We evaluate latency on customized hardware accelerators with the generic paradigm, using E2E-Perf. We report the latency of the searched architecture on three different hardware budgets. }
    \label{tab:imagenet}
    \vspace{-.2cm}
\end{table*}

\subsection{Results on Customized Hardware Accelerators}
\label{sec:darts}
{\noindent \bf Experiment setup.}
We estimate customized hardware accelerators performance of architectures on DARTS~\cite{liu2019darts} search space.
Since the original search space is infeasible~($8^{28}\approx10^{25}$ architectures), we uniformly sampled 1,000K, 200K, 200K architectures to form the training, validation, and test sets. Each architecture is evaluated with E2E-Perf under a relatively large hardware budget~(4800 DSPs 141Mb on-chip memory, comparable to a mid-range cloud processor) for corresponding hardware performance. The dataset collection costs 10 hours in total for each paradigm.

We train our deep hardware loss $\gL_{\tt hw}^{\tt deep}$ using the collected dataset and integrate it into DARTS training pipeline. We consider two paradigms of customized hardware accelerators, namely, pipeline paradigm~(PP) and generic paradigm~(GP). For each paradigm, we train a separate hardware loss. The training time of hardware loss is 2 hours per paradigm. Due to the compact design of the hardware loss model, see~\secref{sec:hw_loss}, the inference time is minimal.
 
DARTS architecture search consists of two stages. In the first stage, we search for the best cell choices based on the validation performance of both classification accuracy and hardware performance. In the second stage, the final searched cells are selected and cells are stacked to construct the final architecture, where the number of cells is 8 for CIFAR10 and 14 for ImageNet. 
Lastly, this final architecture is trained from scratch following DARTS setting to evaluate classification performance. 

For hardware performance~(latency), we prepare three different hardware budgets as hardware constraints: small (1400 DSPs 46Mb on-chip memory), medium (2400 DSPs, 70Mb on-chip memory),  and large (4800 DSPs 141Mb on-chip memory) to cover edge- to cloud-computing. All three budgets feature the same external memory bandwidth with DDR3-1600 and 200MHz working frequency. Note that we search architecture with $\gL_{\tt hw}^{\tt deep}$
trained on large budget hardware performance dataset, and evaluate the searched architecture under three different budgets. This is to validate the generalizability of $\gL_{\tt hw}^{\tt deep}$.

We consider recent baselines including popular NAS and DNAS methods, \eg,
NASNET~\cite{zoph2018learning}, AmoebaNet~\cite{real2019regularized}, ProxylessNAS~\cite{cai2019proxylessnas}, MobileNet V3~\cite{Howard_2019_mobilenetv3,sandler2018mobilenetv2},
SNAS~\cite{xie2018snas}, DARTS~\cite{liu2019darts}, and FBNet~\cite{wu2019fbnet}. Note that for a fair comparison, we adjust FBNet baseline by acquiring a latency look-up-table (LUT) using E2E-Perf with generic paradigm. Note that the LUT-based method can only be applied to the generic paradigm~(GP) as all layers are required to be executed by the same hardware components.

{\noindent \bf Results on CIFAR10.}
We report the quantitative results on CIFAR10 regarding two paradigms, the pipeline paradigm (PP) in ~\tabref{tab:cifar10} and the generic paradigm (GP) in ~\tabref{tab:cifar10_p2}. Each row represents the performance of the architecture searched using the corresponding approach. Note that we evaluate the same architecture on three hardware budgets.

In~\tabref{tab:cifar10}, observe that compared to DARTS, we improve the hardware performance by twice under the medium budget and by 1.7 times under the large budget, as well as improve the classification accuracy by 0.16\%. Our approach reaches the lowest hardware latency among all baselines and comparable classification accuracy. Our approach also features the lowest number of parameters and FLOPS. Notably, all baselines exceed the small budget, indicating their searched architectures are unable to be deployed on such compact hardware with the pipeline paradigm.

In~\tabref{tab:cifar10_p2}, when compared to DARTS, we improve the hardware performance by $1.3\times$ under all three budgets, and improve the classification accuracy by $0.08\%$.
While FBNet LUT only improves hardware performance on large budget. Our approach has the lowest latency, number of parameters, and FLOPS among all baselines, with comparable classification accuracy. 

For both paradigms, we show improved hardware performance on all three hardware budgets, indicating that $\gL_{\tt hw}^{\tt deep}$ learned on the large budget can generalize for other budgets. This significantly reduces the effort of retraining, \ie, given new hardware, it is unnecessary to retrain $\gL_{\tt hw}^{\tt deep}$ as long as they are under the same hardware design paradigm.

\begin{table*}[t]
\setlength{\tabcolsep}{2pt}
\small
\resizebox{\textwidth}{!}{
    \centering
    \begin{tabular}{c|c|ccc|ccc|ccc|ccc}
    \specialrule{.15em}{.05em}{.05em}
     \multicolumn{2}{c|}{Approach} & \multicolumn{3}{c|}{Edge GPU} & \multicolumn{3}{c|}{Edge TPU}& \multicolumn{3}{c|}{Raspi 4}& \multicolumn{3}{c}{Pixel 3}\\
     \hline
    \multirow{2}{4em}{\begin{tabular}{@{}c@{}}Search \\ Algorithm\end{tabular}} &\multirow{2}{4em}{\begin{tabular}{@{}c@{}}Hardware \\ Feedback\end{tabular}}&Latency $\downarrow$& Top1 $\downarrow$ & Params$\downarrow$ &Latency $\downarrow$& Top1$\downarrow$ & Params$\downarrow$ &Latency $\downarrow$& Top1$\downarrow$ & Params$\downarrow$ &Latency $\downarrow$& Top1$\downarrow$ & Params$\downarrow$ \\
       & & (ms) & (\%) & (M)  & (ms) & (\%) & (M)  & (ms) & (\%) & (M)  & (ms) & (\%) & (M)\\
      \hline
      \hline
       & - & \bf3.74 & \bf45.7 & 0.073 & \bf0.60 & \bf45.7 &0.073 & \bf3.84 & \bf45.7 & 0.073 & 1.61 & 45.7 & 0.073\\
      DARTS~\cite{liu2019darts} & FBNet LUT~\cite{wu2019fbnet} & 2.45 & 15.8 &0.073& 0.60 & 45.7 &0.073& 2.96 & 29.1 & 0.073&1.61 & 45.7 & 0.073\\
       & EH-DNAS & \bf2.41 & \bf15.7 & 0.073 & \bf0.51 & \bf29.1 &0.073& \bf2.96 & \bf29.1 & 0.073 & 1.66 &29.1 &0.073\\
      \hline
       & - & \bf6.58 & \bf6.6 & 1.073& \bf1.26 & \bf6.6 & 1.073 & \bf69.19 & \bf6.6 & 1.073 & \bf27.70 & \bf6.6 & 1.073\\
      GDAS~\cite{dong2019search}& FBNet LUT~\cite{wu2019fbnet} & 4.05 & 24.4 & 0.316 & 1.27 & 6.6 & 1.073 & 0.01 & 90.0 & 0.073 & 0.01 & 90.0 & 0.073\\
      & EH-DNAS & \bf1.88 & \bf8.1 & 0.587& \bf1.10 & \bf6.9 & 0.830 & \bf56.89 & \bf6.4 & 0.858 & \bf16.97 & \bf6.4 & 0.858\\
    \specialrule{.15em}{.05em}{.05em}     
    \end{tabular}
}
    \vspace{-.1cm}
    \caption{Quantitative results on CIFAR10 on NAS-Bench-201~\cite{dong2020nasbench201} search space. We show results based on two DNAS algorithms. For each row~(representing an approach), we search for four architectures, each for a type of hardware. We report hardware latency using HW-NAS-Bench~\cite{hwnasbench2021}.}
    \label{tab:201}
    \vspace{-.4cm}
\end{table*}
{\bf\noindent Results on ImageNet.} Following DARTS, we conduct experiments transferring the cells searched on CIFAR10 to ImageNet. We use the cell searched with the generic paradigm and stack 14 layers of cells to construct the final model for ImageNet. 
The generic paradigm allows constructing generic reusable hardware compute units to recurrently process all deep-net layers, which is more widely used in newly developed customized accelerators.
Our training details are provided in the Appendix. 

In~\tabref{tab:imagenet} we report the quantitative results on ImageNet. Compared to DARTS, we improve the classification accuracy by 0.4\% while improving hardware performance by 1.3 times under all three budgets, with 1M reduction in number of parameters. Our approach also achieves the lowest latency, number of parameters, and FLOPs among all baselines.

\subsection{Results on Existing Hardware Processors}\label{sec:201}
{\noindent \bf Experiment setup.}
We use HW-NAS-Bench~\cite{hwnasbench2021} to acquire existing hardware processors~(Edge GPU, Edge TPU, Raspi 4, and Pixel 3) performance on NAS-Bench-201~\cite{dong2020nasbench201} search space. The search space of NAS-Bench-201 contains only 15625 architectures, thus we use all the architectures to train the hardware loss. From HW-NAS-Bench, we obtain the hardware performance of every architecture from NAS-Bench-201 search space. In total, we train four hardware losses, one for each hardware processor. The training time of hardware loss is one hour per hardware. Due to the compact design of hardware loss models, see~\secref{sec:hw_loss}, the inference time is minimal.

During architecture search, we search for the best cell choices based on validation performance on both classification accuracy and hardware performance. We follow all settings in NAS-Bench-201~\cite{hwnasbench2021, dong2021nats} and integrate hardware feedback into two differentiable neural architecture search algorithms~(DARTS and GDAS~\cite{dong2019search}). We consider FBNet~\cite{wu2019fbnet} LUT approach as a baseline and for a fair comparison we resemble their latency look-up table from HW-NAS-Bench.

{\noindent \bf Results on CIFAR10.}
In~\tabref{tab:201}, we report quantitative results on CIFAR10 regarding two search algorithms (DARTS and GDAS) on four hardware processors. With DARTS, our approach improves the hardware performance by an average of $1.3 \times$  while improving the classification accuracy by an average of 20\% for all hardware processors. With GDAS, our approach features an average of $1.9 \times$ hardware performance improvement while maintaining the classification accuracy. 

We note that FBNet LUT does not find meaningful architectures on Raspi 4 and Pixel 3. This is mainly due to the mismatch between real latency and the approximated LUT which assumes additive loss between blocks. We provide more analysis in the next section.
\begin{table*}[t]
    \centering
    \small 
    \begin{tabular}{c|cc|cccc}
    \specialrule{.15em}{.05em}{.05em}
    \multirow{2}{4em}{Approach}&\multicolumn{6}{c}{Average hardware estimation error rate~(\%)$\downarrow$} \\
    \cmidrule{2-7}
    & Accelerator~(GP) & Accelerator~(PP) & Edge GPU & Edge TPU & Raspi 4 & Pixel 3\\
    \hline
    \hline
    FBNet LUT &32.5\textpm0.0 &-&65.3\textpm0.0 & 78.5\textpm0.0 & 879.7\textpm0.0 & 890.5\textpm0.0\\
    EH-DNAS & \bf 3.6\textpm0.5 &\bf 7.9\textpm0.1 & 1.9\textpm0.2 & 4.6\textpm1.6 & 38.0\textpm0.6 & 48.4\textpm7.3\\
    \specialrule{.15em}{.05em}{.05em}     
    \end{tabular}
    \vspace{-0.1cm}
    \caption{Average error rate of approximated hardware performance on different hardware. PP denotes the pipeline paradigm and GP denotes the generic paradigm for customized hardware accelerators. }
    \label{tab:hw_201}
\end{table*}
\begin{figure}[t]
    \centering
    \includegraphics[scale=0.8]{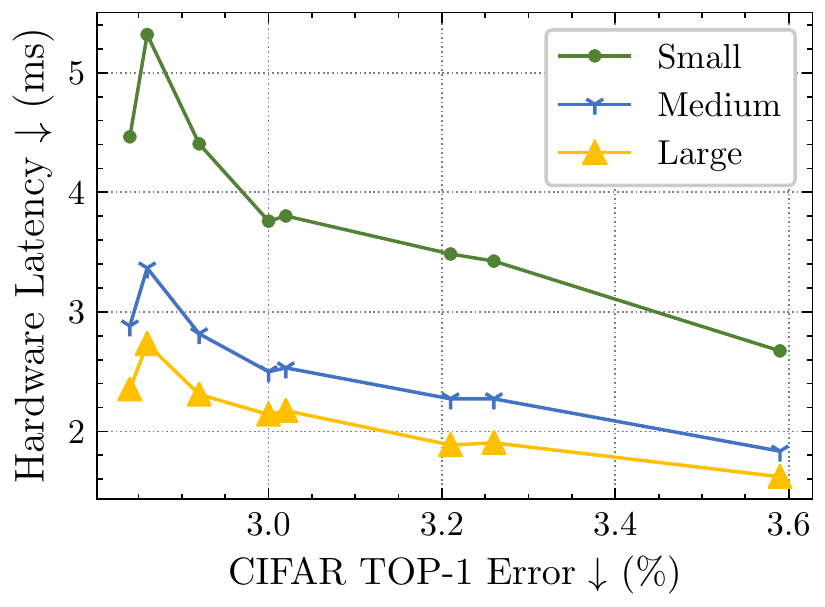}
    \vspace{-0.15cm}
    \caption{Trade-off between hardware performance and classification accuracy. We report EH-DNAS results on three different hardware budgets on customized accelerator of generic paradigm.
    }
    \label{fig:acc_vs_hw}
    \vspace{-0.45cm}
\end{figure} 

\begin{figure*}[t]
\vspace{-0.1cm}
\begin{minipage}{0.5\textwidth}
\centering
 \includegraphics[trim={0 1.3cm 0 .7cm},clip,height=3.0cm]{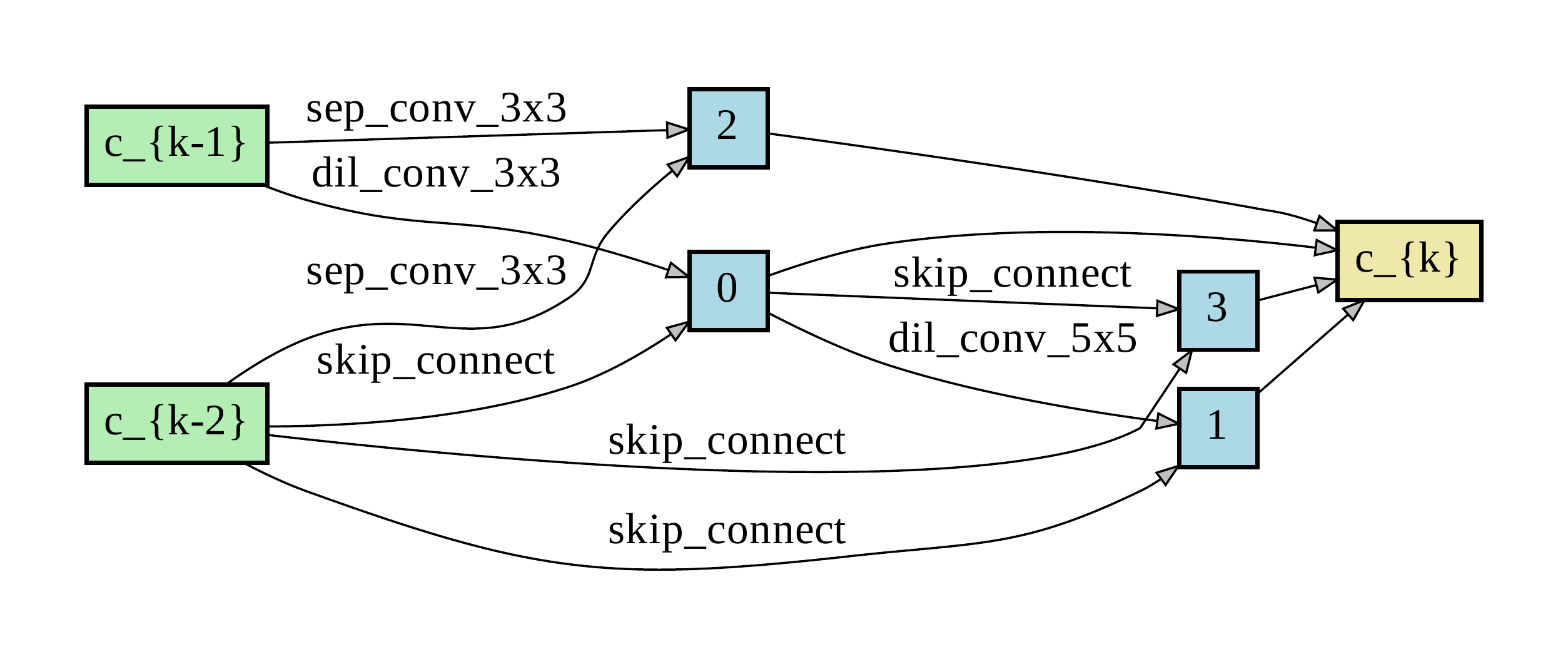}\\
EH-DNAS searched normal cell \\
\vspace{+0.2cm}
\end{minipage}
\begin{minipage}{0.5\textwidth}
\centering
 \includegraphics[trim={0 1.3cm 0 .7cm},clip, height=3.0cm]{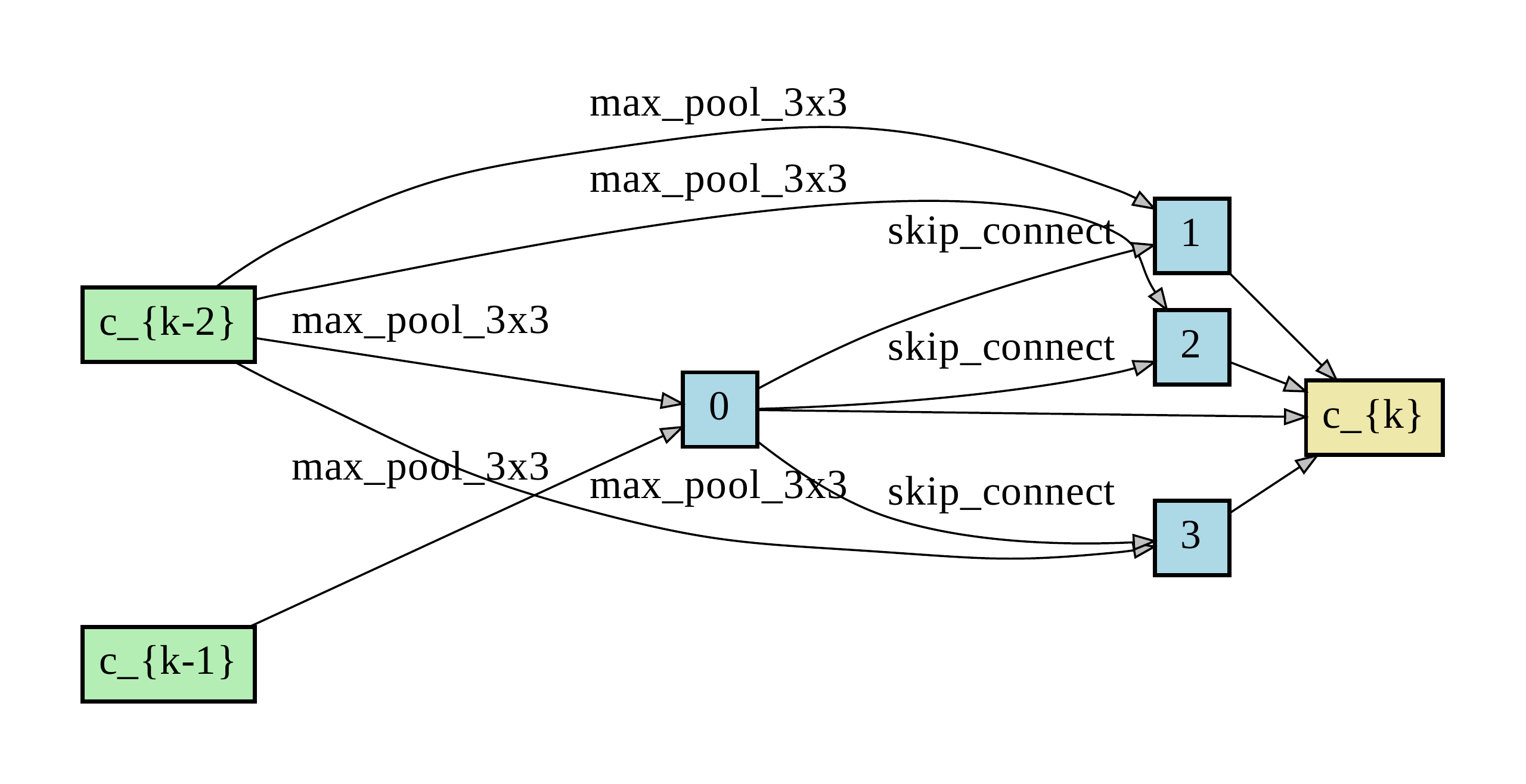}\\
EH-DNAS searched reduction cell\\
\vspace{+0.2cm}
\end{minipage}
\begin{minipage}{0.5\textwidth}
\centering
 \includegraphics[trim={0 1.3cm 0 .9cm},clip,height=4cm]{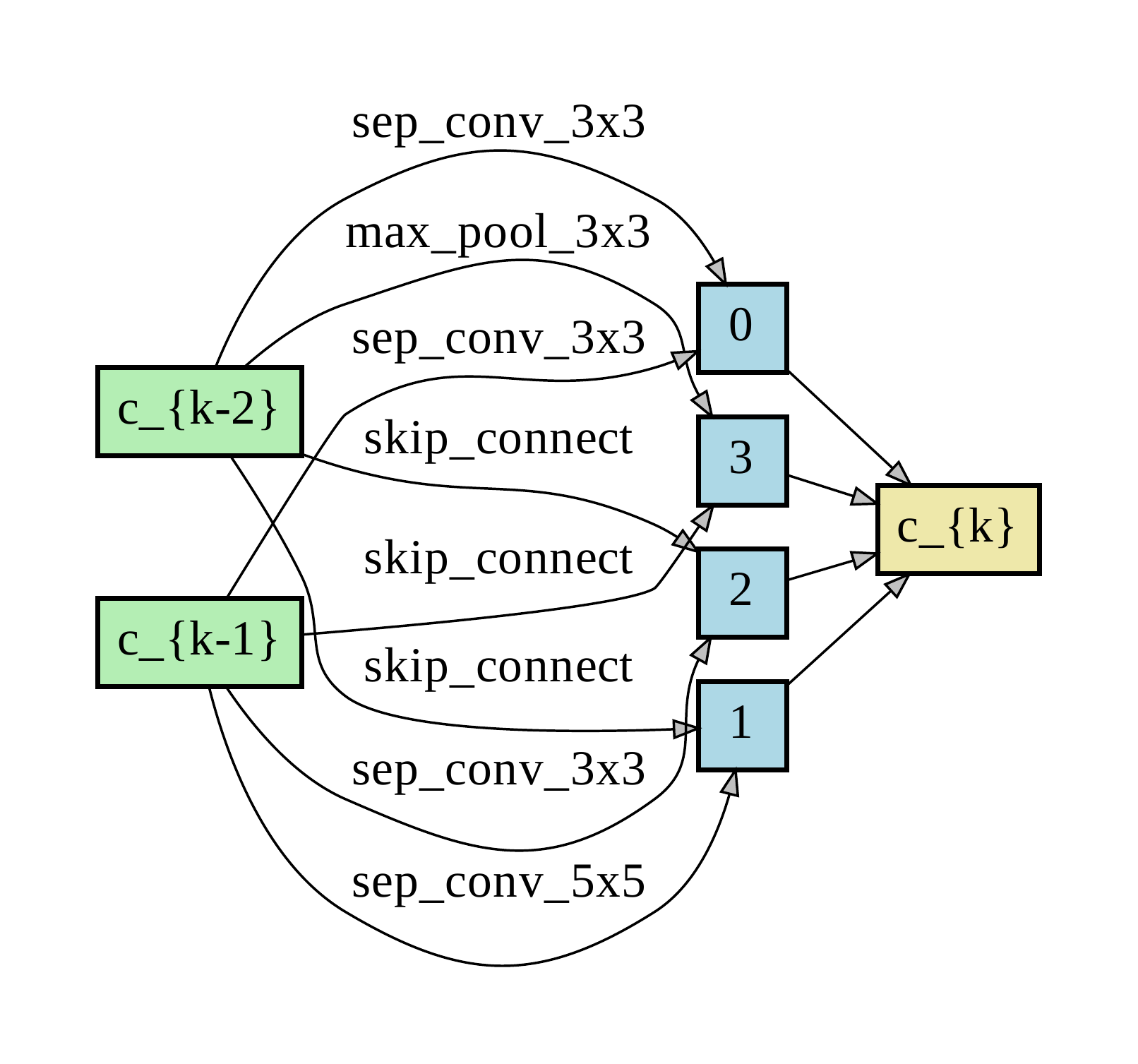}\\
FBNet LUT searched normal cell\\
\end{minipage}
\begin{minipage}{0.5\textwidth}
\centering
 \includegraphics[trim={0.5cm 1.3cm 0.5cm .9cm},clip,height=2.5cm]{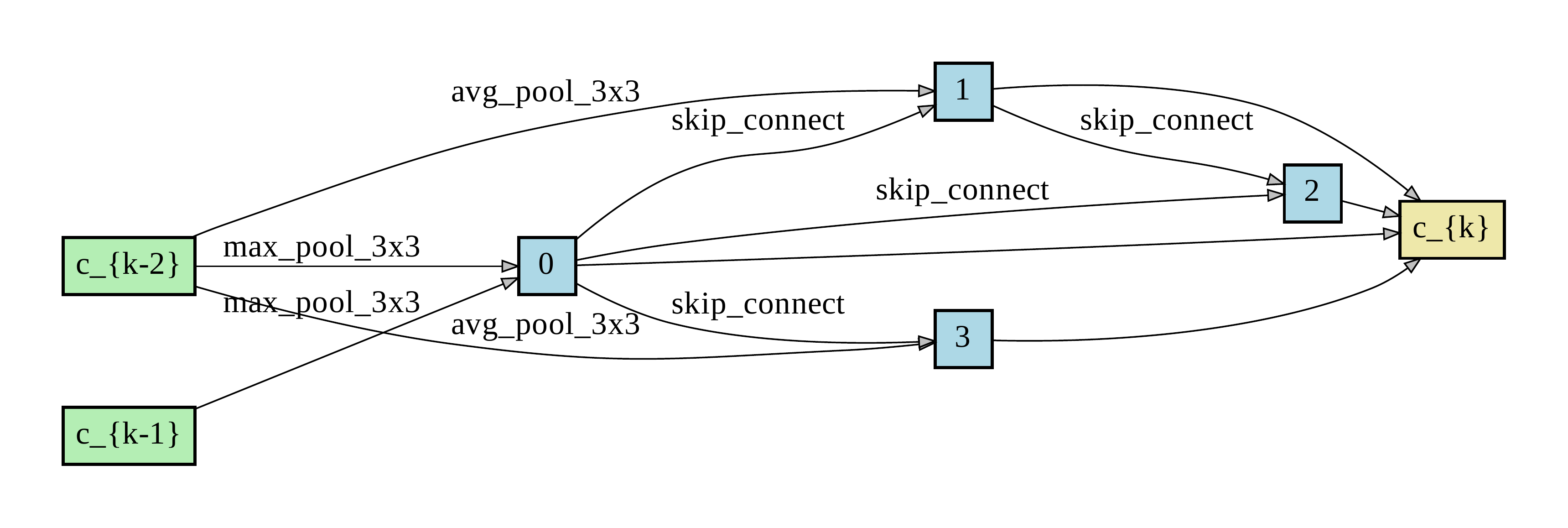}\\
FBNet LUT searched reduction cell\\
\vspace{-1.5cm}
\end{minipage}
\caption{Searched final cells on CIFAR10.} %
\label{fig:cells}
\vspace{-0.3cm}
\end{figure*}
\subsection{Analysis}
\label{sec:analysis}
{\noindent \bf Hardware performance estimation.} To better understand the mechanism behind our approach, we analyze the hardware estimation ability. We obtain the latency LUT for different hardware for FBNet LUT approach. The overall network latency is calculated by summing up the latency from each layer. Note that FBNet LUT only applys to GP. 

In~\tabref{tab:hw_201}, we report the relative average error rate (\%) of hardware performance prediction. The error rate is defined as the absolute difference between predicted and true hardware performance, divided by true hardware performance. The true hardware performance are measured by E2E-Perf for customized accelerators and HW-NAS-Bench for existing hardware processors.
We report mean and standard deviation over three runs with different random initialization seeds. Note that the LUT-based approach is deterministic.

Observe that the LUT-based approach leads to more significant prediction errors for Raspi 4 and Pixel 3, where mobile CPUs are involved. This could lead to the failure cases of the LUT approach in~\tabref{tab:201}. The major reason for this phenomenon is that mobile CPUs perform not only deep-net inference but also run other tasks, such as the operating systems. In addition, the limited memory access bandwidth in mobile CPU is likely to be the bottleneck that significantly slows down the overall performance. In comparison, our approach reaches much lower average error rates for all hardware platforms, as well as accommodate both pipeline and generic paradigms.

{\noindent \bf Deep hardware loss's model complexity.} We examine how the model complexity of our deep hardware loss $\cL_{\tt}^{\tt deep}$ influences the hardware performance estimation. We observe that the estimation error rate increases by 2\% with embedding size 50 and increases by 9\% with embedding size 2. Increased complexity does not necessarily lead to better estimation. Our final choice of embedding size 10 leads to an error rate of 3.6\% as reported in~\tabref{tab:hw_201}. We aim for a model with minimum complexity that is sufficient for accurate hardware performance estimation, and proper selection of model complexity is important.

{\noindent \bf Hardware and classification performance trade-off.} In~\figref{fig:acc_vs_hw}, we show the trade-off between classification and hardware performance.
We obtain different searched architectures by adjusting the hyperparameter $\beta$ that controls the scale of the hardware loss term. Larger $\beta$ generally leads to better hardware performance.
The experiments are conducted on customized accelerators with the generic paradigm. Note that with a slight compromise of classification accuracy, \ie, less than 1\%, we can improve hardware performance by almost twice. This shows the benefit of optimizing hardware-aware metrics.
With a proper $\beta$, we can find the architecture that meets the classification accuracy requirement with optimized hardware performance.

{\noindent \bf Searched cells.} In~\figref{fig:cells}, we show the cells found on DARTS search space. Observe that FBNet LUT tends to simply reduce the layer complexity to achieve better hardware performance, while our approach is able to search for more complex architectures. This potentially explains that LUT-based approach could benefit hardware efficiency yet the simplicity of its approximation limits the effectiveness.

\section{Conclusion}\label{sec:conc}
We present EH-DNAS, an end-to-end hardware-aware DNAS framework. We integrate hardware performance benchmarking, differentiable hardware loss approximation, and DNAS to search for efficient and accurate architectures. We also propose E2E-Perf, an accurate benchmarking tool for customized hardware accelerators. On CIFAR10 and ImageNet, EH-DNAS improves the hardware performance by an average of $1.4\times$ on customized accelerators and $1.6\times$ on existing hardware processors while maintaining the classification accuracy.

\clearpage

{\small
\bibliographystyle{ieee_fullname}
\bibliography{egbib}
}

\end{document}